\documentstyle[11pt]{article}
\title{Instructions for Temporal Annotation of Scheduling
Dialogs\footnote{This research was supported in part by the US Department of 
Defense under contract number \mbox{MDA904-96-C-0354}.}}
\author{Tom O'Hara, Janyce Wiebe, and  Karen Payne}
\begin{document}
\maketitle

\begin{abstract}

Human annotation of natural language facilitates standardized
evaluation of natural language processing systems and supports
automated feature extraction. This document consists of instructions
for annotating the temporal information in scheduling dialogs, dialogs
in which the participants schedule a meeting with one another.
Task-oriented dialogs, such as these are, would arise in many useful
applications, for instance, automated information providers and
automated phone operators.  Explicit instructions support good
inter-rater reliability and serve as documentation for the classes
being annotated.

\end{abstract}

\newcounter{examplenum}[part]                 
\newcounter{nextexamplenum}[part]             

\def\newexample{\addtocounter{examplenum}{1}} 
\def\exref{\theexamplenum}                    

\def\exrefnext{\setcounter{nextexamplenum}{\value{examplenum}}\addtocounter{nextexamplenum}{1}\thenextexamplenum}
\def\exrefnextN#1{\setcounter{nextexamplenum}{\value{examplenum}}\addtocounter{nextexamplenum}{#1}\thenextexamplenum}

\def\oldexample {\newexample\begin{quote}\begin{tabular}{ll}}
\def\endoldexample {\end{tabular}\end{quote}}

\def\example {\newexample\begin{quote}\begin{tabular}{llll}}
\def\endexample {\end{tabular}\end{quote}}


\section{Introduction}

Research systems in computational linguistics are no longer assessed solely by
appeals to researcher intuition on selected examples.
Instead, system results usually are compared to
independent interpretations of naturally occurring text. 
Therefore, the main
purpose of this temporal coding task is to provide a benchmark for evaluating
the performance of
Artwork\footnote{Artwork is being developed at NMSU's Computing Research
Laboratory as part of a joint project with CMU on investigating ways to
facilitate machine translation of scheduling dialogs.}
on tracking the times discussed in scheduling
dialogs. This includes the inference of missing or underspecified temporal
information. To facilitate the evaluation, the temporal information is being
recorded in a standardized format.

This coding can also serve as a starting point for obtaining a tagged
corpus of temporal expressions in naturally occurring discourse. Such
a corpus makes it easier to compare the performance of different
research systems on the same problem. It also permits for statistical
analysis, which might be useful for extracting temporal features.

Each dialog to be coded concerns two people trying to arrange for a meeting.
For each utterance, you are to specify the time that is being referred to
either explicitly or implicitly. This should only be based on an
interpretation of the utterances that have already been encountered (the
dialog context). That is, you should not revise answers in light of subsequent
utterances.

Furthermore, you should interpret the text based on your common-sense
intuitions regarding meetings, such as that they usually occur during normal
working hours.  However, in order to restrict the wide range of plausible
inferences that can be applied, you need to limit such inferences to what can
be {\em readily inferred} from the dialog. This vague notion will be discussed
in detail later. For now, consider it as covering the obvious
inferences. Surprisingly, these are often the most difficult to model via a
computer.

See the guidelines section below for more information about these and other
requirements. First, an example dialog will be discussed to show what is
expected of you. Then, the coding representation is explained. You need to
adhere to this format to ensure that the results are recorded in a common
format. After the detailed list of guidelines, there is a section with
additional coding advice. The appendix gives a complete example of a dialog
coded based on these instructions.

\section{Background}

The following transcript illustrates the type of scheduling dialog
you will be tagging. It is given in both the original Spanish and in
the English translation. You do not need to refer to the Spanish
version to perform this task; it is provided as background material
for those who can read Spanish.\footnote{Subtle distinctions are
occasionally lost in the translation, but this usually doesn't affect the
temporal information.}

\begin{quote}

S1: (?Quihubo, Primo, qu\'{e} pas\'{o}?)
\newline
{\bf (1) What's up, Primo, what's happening?}

S2: (Nada, aqu\'{i} nada m\'{a}s, ?c\'{o}mo has estado?)
\newline
{\bf (2) Not much, same old thing, how you been?}

S1: (Muy bien. Oye, necesito juntarme contigo. Necesitamos, este,
discutir eh, uh, algo del proyecto. ?Cu\'{a}ndo tienes tiempo?)
\newline
{\bf (3) Good. (4) Listen, I need to talk to you. (5) We need to, ahh, 
talk about, ahh, uh, something about the project. (6) When do you have time?}

S2: H\'{i}jola, pos, que d\'{i}a estas pensando en...
\newline
{\bf (7) Jeez, well, what day were you thinking about...}

S1: Pos si se puede ma\~{n}ana. Si se puede ma\~{n}ana. ?Tienes chance ma\~{n}ana?
\newline
{\bf (8) Well tomorrow, if you can. (9) Tomorrow if you can. 
(10) Do you have a chance tomorrow?}

S2: (Pos depende en qu\'{e} hora porque mira, para empezar como horita de
las ocho a las diez, tengo una cita con el doctor y ya sabes como le
entretiene a uno.)
\newline
{\bf (11) Well, it depends on what time because, look, to begin with I have a
doctor's appointment between eight and ten and you know the runaround
they give you.}

S1: (S\'{i})
\newline
{\bf (12) Yeah.}

\end{quote}

The dialogs are divided into groups of consecutive utterances by the
same speaker, which are called {\em turns}. Most of the turns have
just one or two utterances, but a few have several, especially if the
speaker makes a correction.

In this dialog, the first two turns are typical greetings that the
speakers exchange.  No temporal information has been conveyed. Next,
speaker 1 introduces the need to have a meeting and then asks the
other person to indicate availability. In reply, speaker 2 asks for
a clarification. Although there is mention of ``when'' and ``what
day'', these are not specific times, so they are not
represented. Therefore, the entries for the first seven utterances
would indicate no temporal information:

\begin{quote}
\begin{tabular}{|lc|c|c|c|c|c|}
\hline
\# & Start & WeekDay & Month & Date  & Hour & Time of Day \\
   & End & WeekDay & Month & Date  & Hour & Time of Day \\
\hline
1 & & null & null & null & null & null \\
  & & null & null & null & null & null \\
\hline
2 & & null & null & null & null & null \\
  & & null & null & null & null & null \\
\hline
3 & & null & null & null & null & null \\
  & & null & null & null & null & null \\
\hline
4 & & null & null & null & null & null \\
  & & null & null & null & null & null \\
\hline
5 & & null & null & null & null & null \\
  & & null & null & null & null & null \\
\hline
6 & & null & null & null & null & null \\
  & & null & null & null & null & null \\
\hline
7 & & null & null & null & null & null \\
  & & null & null & null & null & null \\
\hline
\end{tabular}
\end{quote}

This shows that time is being represented by intervals rather than by
points, as there are starting and ending entries for all fields.\footnote{An
alternative representation would be to encode the start and end points
separately along with the relation between the points. For instance, the
entry for (8) would be split into two entries with a {\em precedes}
relation between the two.}
The interval-based
approach is commonly used in natural language processing, since it is
closer to the way people usually express time periods. The example
also shows that all the fields are explicitly filled in: in cases
where no information is known, the special entry ``null'' is
used. This is designed to eliminate errors due to coder omissions.

Starting with utterance (8), specific times are being discussed.
Speaker 1 suggests ``tomorrow'' for a potential meeting. Since the
dialog date is Tuesday, May 11th, 1993, we know that the time in
question is Wednesday, May 12th. The next two utterances reiterate
this request, so the times will be the same:

\begin{quote}
\begin{tabular}{|lc|c|c|c|c|c|}
\hline
\#  & Start & WeekDay & Month & Date & Hour & Time of Day \\
    &   End & WeekDay & Month & Date & Hour & Time of Day \\
\hline
8 & & Wednesday & May & 12 & null & null \\
  & & Wednesday & May & 12 & null & null \\
\hline
9 & & Wednesday & May & 12 & null & null \\ 
  & & Wednesday & May & 12 & null & null \\
\hline
10 & & Wednesday & May & 12 & null & null \\
   & & Wednesday & May & 12 & null & null \\
\hline
\end{tabular}
\end{quote}

It might seem odd that both the starting and ending times are the
same. Doesn't this mean the interval has no length? No, since the hour
and time of day fields are left unspecified. The representation
indicates that the period in question starts some time on May 12th and
ends some time on the same day. However, there are cases when the
interval length will be zero. For instance, (10') ``But remember the
office closes tomorrow at 3pm'' would be represented as

\begin{quote}
\begin{tabular}{|lc|c|c|c|c|c|}
\hline
10' & & Wednesday & May & 12 & 3:00 & afternoon \\
    & & Wednesday & May & 12 & 3:00 & afternoon \\
\hline
\end{tabular}
\end{quote}

In response to the suggested date of Wednesday, May 12th, the other
person gives a constraint that the meeting can't occur between eight
and ten due to a doctor's appointment. Note that there is no mention
of the meeting date, but this is readily inferred from the dialog
context. Finally, the last utterance conveys no temporal information.

\begin{quote}
\begin{tabular}{|lc|c|c|c|c|c|}
\hline
11 & & Wednesday & May & 12 & 8:00 & morning \\
   & & Wednesday & May & 12 & 10:00 & morning \\
\hline
12 & & null & null & null & null & null \\
   & & null & null & null & null & null \\
\hline
\end{tabular}
\end{quote}

As indicated in the table, we also inferred that the time for (11) is in the
morning.  Although it might seem obvious that it must be in the morning, you
have to be careful when you make this type of judgement. In this case, you'd
be surprised if the appointment were in the evening (especially since it's a
doctor's appointment). Thus, when in doubt about a time, determine whether it
would surprise you if it weren't the case. Just leave the field as `null',
otherwise.

\section{Code Format}

For each distinct time mentioned in the dialog, please fill out
the following information, using 'null' to indicate information 
that cannot be inferred and commas to separate fields. A template
will be provided to help ensure that the correct format is applied.
\begin{quote}
\begin{tabular}{llllll}
[Label, & SWeekDay, & SMonth, & SDate, & SHourSpec, & STimeOfDay, \\
& EWeekDay, & EMonth, & EDate, & EHourSpec, & ETimeOfDay]
\end{tabular}
\end{quote}
The {\bf S} prefixes fields for the starting time and the {\bf E}
prefixes those for the ending time.
Each of the components in this structure is given in the following
format, where the qualifiers are optional:
\begin{center}
([Qualifiers] TimeComponent) \\
\end{center}
Example ({\exrefnext}) shows the exact format that is required.
Possible values for the components fields are listed in figure
\ref{component-values}.
\begin{example}
        ({\exref}) & & & ``I'm free late in the afternoon on Monday the 19th 
                           until Wednesday'' \\
 \\
        & & [{\exref}, & (monday), (august), (19), (null), (late, afternoon), \\
        & &            & (wednesday), (august), (21), (null), (null)] \\
\end{example}

\begin{figure}
\begin{tabular}{|l|l|}
\hline
Field & Description \\
\hline
Label & Numeric label of the utterance (e.g., '1', '19a') \\
      & Use '1\_alt1', etc. for alternatives (see below) \\
\hline
WeekDay & sunday, monday, ..., saturday \\
\hline
Month & january, february, ..., december \\
\hline
Date & 1, 2, 3, ..., 31 \\
\hline
HourSpec & 10, 12:15 \\
         & \\
         & Omit the AM or PM indicator, \\
         & since that is covered by the Time of Day field. \\
\hline
TimeOfDay & morning, afternoon, evening, \\
          & breakfast, lunch, dinner, \\
          & all-day \\
          & \\
          & Since meetings could last an entire day, be careful not \\
          & to assume it will just be in the morning or afternoon. \\
\hline
\end{tabular}

\begin{tabular}{ll}
\end{tabular}

\begin{tabular}{|l|l|}
\hline
Qualifiers & Description \\
\hline
before, after, during, & Mainly for use with TimeOfDay ('before lunch'), \\
early, mid, late & but could also be used with other fields ('after the fifth'). \\
\hline
\end{tabular}
\caption{Possible Values for the Components of the Temporal Representation}
\label{component-values}
\end{figure}

A few examples will be given just to illustrate the desired
format. The rationale for the content of the representation is covered
in the next section.
First of all, since qualifiers can apply to any field, the value
for each component is actually a list of items.
These lists are enclosed in parentheses to eliminate confusion.
The whole structure is then enclosed in square brackets.

\begin{example}
        ({\exref}) & & s1: & ``Are you free Wednesday, September 4th'' \\
        & & [{\exref}, & (wednesday), (september), (4), (null), (null), \\
        & &            & (wednesday), (september), (4), (null), (null)] \\
\end{example}

\begin{example}
        ({\exref}) & & s2: & ``Yes but only early in the morning'' \\
        & & [{\exref}, & (wednesday), (september), (4), (null), (early, morning), \\
        & &            & (wednesday), (september), (4), (null), (early, morning)] \\
\end{example}

\begin{example}
        ({\exref}) & & s1: & ``Oh, then how about after the fifth?'' \\
        & & [{\exref}, & (null), (september), (after, 5), (null), (null), \\
        & &            & (null), (null), (null), (null), (null)] \\
\end{example}

\section{Coding Guidelines}

Please keep in mind the following guidelines for specifying the times. The
examples given assume the dialog date is Monday 19 August 1996. The calendar 
for this period follows:

{\small
\begin{quote}

\begin{tabular}{|r|r|r|r|r|r|r|}
\multicolumn{7}{c}{August 1996}\\
\hline
\mbox{ Sun } & \mbox{ Mon } & \mbox{ Tues} & \mbox{ Wed } & \mbox{Thurs} & \mbox{ Fri } & \mbox{ Sat } \\
\hline
   &    &    &    &  1 &  2 &  3 \\
\hline
  4 &  5 &  6 &  7 &  8 &  9 & 10 \\
\hline
 11 & 12 & 13 & 14 & 15 & 16 & 17 \\
\hline
 18 & {\bf 19} & 20 & 21 & 22 & 23 & 24 \\
\hline
 25 & 26 & 27 & 28 & 29 & 30 & 31 \\
\hline
\end{tabular}
\end{quote}
}

In what follows, the labeling convention for these examples uses a
number followed by an optional letter suffix (e.g., `1a'). The number
is based on the example number, rather than the utterance number. The
letter is used when there is more than one utterance for the same
example.  Since there might be several examples together, this might
be a little confusing. However, two utterances will be related only if
the labels use the same example number as a prefix. For instance, in
the following, utterances ({\exrefnext}b) and ({\exrefnextN{2}}a)
would be unrelated.

\begin{example}
        ({\exref}a) & & s1: & ``Good morning, Peter.'' \\
        ({\exref}b) & & s2: & ``Good morning, and how are you, Mary?'' \\
\end{example}

\begin{example}
        ({\exref}a) & & s1: & ``What day is best for you?'' \\
        ({\exref}b) & & s2: & ``None really, since my schedule's quite full.'' \\
\end{example}

The guidelines follow.

\begin{enumerate}

\item
{\em Fill in missing temporal information based on what can be readily
inferred from the context of the dialog.} Avoid relying too much on 
common-sense assumptions such as about the most likely times for meetings.
Unfortunately, there are no clear boundaries. Instead, you must use 
your intuition about what seem to be reasonable assumptions.

\begin{example}
        ({\exref}a) & & s1: & ``Let's meet tomorrow morning''. \\
        ({\exref}b) & & s2: & ``OK, how about 8 o'clock'' \\
 \\
        {\small ok}  & & [{\exref}a, & (tuesday), (august), (20), (null), (morning), \\
        &            & & (tuesday), (august), (20), (null), (null)] \\
        & & [{\exref}b, & (tuesday), (august), (20), (8), (morning), \\
        & &            & (tuesday), (august), (20), (null), (null))] \\
\end{example}

\begin{example}
        ({\exref}a) & & s1: & ``We need to discuss the project soon.'' \\
        ({\exref}b) & & s2: & ``Well, let's meet Thursday after lunch.'' \\
 \\
        {ok} & & & \\
        & & [{\exref}a, & (null), (null), (null), (null), (null), \\
        &           & & (null), (null), (null), (null), (null)] \\
        & & [{\exref}b, & (thursday), (august), (22), (null), (after, lunch), \\
        & &             & (thursday), (august), (22), (null), (null)] \\
 \\
\end{example}

\begin{example}
        ({\exref}a) & & s1: & ``Let's meet Friday afternoon''. \\
        ({\exref}b) & & s2: & ``OK, how about 3pm''. \\
 \\
        {\bf not ok} & & & \\
        & & [{\exref}a, & (friday), (august), (23), (null), (afternoon), \\
        &               & & (friday), (august), (23), (null), (null)] \\
        & & [{\exref}b, & (friday), (august), (23), (3), (afternoon), \\
        & &             & (friday), (august), (23), (null), (afternoon)] \\
\end{example}

The representation for ({\exref}b) suggests the meeting ends in the 
afternoon (presumably since people usually don't stay late on Fridays). Thus
it is a case of extrapolating too much from the dialog. Note that 'evening'
would also be inappropriate, since the meeting might very well be short.
Therefore, it's better to be noncommittal and use 'null'.

\item
{\em Only consider the previous utterances when interpreting the time
of an utterance.} So don't read ahead if you are unsure of the
time. Likewise, don't revise previous answers. There might be
transcription errors in the dialogs, which were recorded from actual
conversations. Even if a later utterance reveals an error in an
earlier one, do {\bf not} go back and revise it.

Note that you may not be able to determine what all of the times
are---sometimes speakers are too vague for a definitive time to be determined.
Fill in a time when you are reasonably certain what the speakers mean.

\item
{\em Consider weeks as just workweeks.} Therefore, omit weekends
unless the dialog suggests otherwise.

\begin{example}
        ({\exref}) & & s1: & ``I am free all this week'' \\
        & & ({\exref}, & (monday), (august), (19), (null), (null) \\
        & &            & (friday), (august), (23), (null), (null)) \\
\end{example}

When specifiers, such as first or last, apply to partial workweeks,
the interpretation can be quite subjective. A rule of thumb is to
consider only those with more than two working days.

\begin{example}
        ({\exref}) & & s1: & ``I was in Portland the first week of August'' \\
        & & ({\exref}, & (monday), (august), (5), (null), (null) \\
        & &            & (friday), (august), (9), (null), (null)) \\
\end{example}

This shows that the period discussed is the interval from Monday the
fifth to Friday the ninth, treated as a single unit. (The partial
workweek Aug 1-2 is skipped.)

\item
{\em Be as specific as possible about the time for a meeting or for the
speaker's availability.} For instance, if a day of the week is
specified relative to some week, just code the time for the day (not
the entire week), as illustrated in ({\exrefnext}a).

\begin{example}
        ({\exref}a) & & s1: & ``Let's meet Friday next week'' \\
        ({\exref}b) & & s2: & ``I can only meet next month'' \\
        ({\exref}c) & & s1: & ``OK, then how about the first week'' \\
 \\
        & & ({\exref}a, & (friday), (august), (30), (null), (null), \\
        & &            & (friday), (august), (30), (null), (null)) \\
        & & ({\exref}b, & (sunday), (september), (1), (null), (null), \\
        & &            & (monday), (september), (30), (null), (null)) \\
        & & ({\exref}c, & (monday), (september), (2), (null), (null), \\
        & &            & (friday), (september), (6), (null), (null)) \\
\end{example}

({\exref}b) shows that months should be interpreted as ranges from the
first of the month to the last. Weeks are handled similarly except
that weekends are not included.

\item
{\em If no end time is mentioned, then code the end time using as much of 
the start time as possible.} For instance, if the start time mentions 
the day and the exact starting time in hours and minutes, use the
day in the end time as well, but omit the HourSpec:
\begin{example}
        ({\exref}) & & s1: & ``Let's meet Thursday at 9am'' \\
\\
        & & ({\exref}, & (thursday), (august), (22), (9), (morning), \\
        & &            & (thursday), (august), (22), (null), (null)) \\
\end{example}

\item
{\em Don't just encode the time as literally given.} Instead, encode it
based on an interpretation giving the most information regarding the
meeting or other scheduling event. For example, information on an
entire meeting is more informative than just information on its start
or end.

\begin{example}
        ({\exref}a) & & s1: & ``The meeting has to be Friday at 8am.'' \\
        ({\exref}b) & & s2: & ``OK, just so it ends at 10am.'' \\
\\
        & & ({\exref}a, & (friday), (august), (23), (8), (morning), \\
        & &             & (friday), (august), (23), (null), (null)) \\
        & & ({\exref}b, & (friday), (august), (23), (8), (morning), \\
        & &             & (friday), (august), (23), (10), (morning)) \\
\end{example}

Taken literally, ({\exref}a) would be represented as a point. But
since meetings take time, the ending hour is left
unspecified. Similarly, the encoding for ({\exref}b) provides more
information about the meeting than what a literal interpretation of
``at 10am'' would provide.

\item
{\em Interpret simple events as instantaneous unless 
such events are usually
considered as part of an extended event.} As above, the encoding should
indicate the time most relevant for the scheduling domain.

\begin{example}
        ({\exref}a) & & s1: & ``Let's meet tomorrow afternoon at 3pm'' \\
        ({\exref}b) & & s2: & ``Would we then adjourn at 5pm?'' \\
        ({\exref}c) & & s1: & ``Yes, I'll remind you at 1pm.'' \\
\\
        & & ({\exref}a, & (tuesday), (august), (20), (3), (afternoon), \\
        & &             & (tuesday), (august), (20), (null), (null)) \\
        & & ({\exref}b, & (tuesday), (august), (20), (3), (afternoon), \\
        & &             & (tuesday), (august), (20), (5), (afternoon)) \\
        & & ({\exref}c, & (tuesday), (august), (20), (1), (afternoon), \\
        & &             & (tuesday), (august), (20), (1), (afternoon)) \\
\end{example}

It is natural to consider ({\exref}a) as referring to a point in time,
since you might interpret ``meet'' as ``encounter'', an instantaneous
occurrence or simple event. But since we're concerned primarily with
the scheduling domain, you should generally interpret ``meet'' as
``hold meeting'', a prolonged occurrence or complex event. Similarly,
({\exref}b) might be taken to refer to an instantaneous occurrence,
just considering the adjournment itself. But it also should be
represented as an interval, because the simple adjournment event is
part of the complex meeting event (the focus of the domain). 
However, ({\exref}c) is represented by a point, since reminders are
generally independent simple events.

Note that the distinction between instantaneous and prolonged occurrences 
(or simple versus complex event) is easier to see in other domains:
\begin{example}
        ({\exref}a) & & s1: & ``John walked the dog at 6pm'' \\
        ({\exref}b) & & s2: & ``Spot bit John at 6:15'' \\
\\
        & & ({\exref}a, & (null), (null), (null), (6), (evening), \\
        & &             & (null), (null), (null), (null), (evening)) \\
        & & ({\exref}b, & (null), (null), (null), (6:15), (evening), \\
        & &             & (null), (null), (null), (6:15), (evening)) \\
\end{example}

({\exref}a) is represented as an interval since walking a dog is a
complex event. ({\exref}b), in contrast, is represented by a point,
since a dog-bite event usually is (relatively) instantaneous. 

\item
{\em Consider the ``big picture'' when deciding on event
interpretations.} This guideline is implicit in the previous two, but
it is important enough to emphasize separately. Since the same event
can be looked at from different viewpoints (the event {\em aspect}),
there might be confusion on what to encode. In ({\exrefnext}) below,
the entire event is being referred to, whereas in ({\exrefnext}'),
just the start. Again, you should consider the temporal references at
the level of the entire meeting event, instead of at the level of the
viewpoint. Thus, ({\exrefnext}) and ({\exrefnext}') are represented in
the same way.
\begin{example}
        ({\exref}) & & s1: & ``Let's meet in the afternoon'' \\
        & & ({\exref}, & (null), (null), (null), (null), (afternoon) \\
        & &            & (null), (null), (null), (null), (null)) \\
\\
        ({\exref}') & & s2: & ``Let's start the meeting in the afternoon'' \\
        & & ({\exref}', & (null), (null), (null), (null), (afternoon) \\
        & &             & (null), (null), (null), (null), (null)) \\
\end{example}

Some people may read sentences such as ({\exref}) as meaning that the
entire meeting must be held in the afternoon.  For consistency of
tagging, however, please encode these (and similar) examples as
specified in this document.

\item
{\em Treat discrete times individually.} Please, label each alternative
(in the disjunction) with the suffix '\_alt1', '\_alt2', etc., as illustrated. 
Similarly, label each ``required'' time (in a conjunction) with '\_and1',
'\_and2', etc.

\begin{example}
        ({\exref}) & & s1: & ``I can meet Wednesday or Friday'' \\
        & & ({\exref}\_alt1, & (wednesday), (august), (21), (null), (null), \\
        & &                  & (wednesday), (august), (21), (null), (null)) \\
        & & ({\exref}\_alt2, & (friday), (august), (23), (null), (null), \\
        & &                  & (friday), (august), (23), (null), (null)) \\
\end{example}

\begin{example}
        ({\exref}) & & s1: & ``I can meet Tuesday and Thursday'' \\
        & & ({\exref}\_and1, & (tuesday), (august), (21), (null), (null), \\
        & &                  & (tuesday), (august), (21), (null), (null)) \\
        & & ({\exref}\_and2, & (thursday), (august), (22), (null), (null), \\
        & &                  & (thursday), (august), (22), (null), (null)) \\
\end{example}

\item
{\em Treat time ranges as a unit represented by one interval.}
That is, use a single descriptor instead of a series of alternatives.

\begin{example}
        ({\exref}) & & s1: & ``I will be away the 22nd through the 26th'' \\
        & & ({\exref}, & (thursday), (august), (22), (null), (null), \\
        & &            & (monday), (august), (26), (null), (null)) \\
\end{example}

\item
{\em When in doubt about an assumption you are making, ask yourself whether you
would be surprised it if weren't the case.}

\begin{example}
        ({\exref}) & & s1: & ``Let's meet at two'' \\
        & & ({\exref}, & (null), (null), (null), (2), (afternoon) \\
        & &            & (null), (null), (null), (null), (null)) \\
\end{example}

The meeting of ({\exref}) is assumed to be in the afternoon, since you
would probably be surprised if the meeting were instead at 2am. (You
might object that you wouldn't be surprised at a 2am clandestine
meeting, but remember to assume normal working hours and activities.)

\begin{example}
        ({\exref}) & & s1: & ``Let's meet at eight'' \\
        & & ({\exref}, & (null), (null), (null), (8), (null)) \\
        & &            & (null), (null), (null), (null), (null)) \\
\end{example}

Here you probably wouldn't be surprised if the meeting was at 8pm,
so coding the meeting as in the morning might be assuming too much
(unless, of course, something in the prior context suggests 
that the morning is intended).

\item
{\em Fill in fields that can be determined from other fields or from the
dialog date.} For instance, resolve dates to days of the week and use
the current month when no other month could reasonably apply.
Likewise, resolve temporal adverbs and phrases (e.g., ``tomorrow'',
``in two days'') to absolute dates based on the dialog date.

\begin{example}
        ({\exref}) & & s1: & ``Today is the nineteenth right?'' \\
        & & ({\exref}, & (monday), (august), (19), (null), (null) \\
        & &            & (monday), (august), (19), (null), (null)) \\
\end{example}

\begin{example}
        ({\exref}) & & s1: & ``Are you free tomorrow morning?'' \\
        & & ({\exref}, & (tuesday), (august), (20), (null), (morning) \\
        & &            & (tuesday), (august), (20), (null), (null)) \\
\end{example}

\item
{\em When trying to decide how much common-sense reasoning you can apply to
the interpretation, consider how ``specific'' the times are that have
been mentioned.} In regards to time, it is generally safer to make
assumptions going from the specific to the general.  So it is safe to
assume a meeting that starts at 1pm ends on the same day. But it is
risky to assume that a meeting on a Monday won't start before 9am.

Similarly, consider the range of possibilities and the likelihood
of the alternatives. If someone says ``at 3 o'clock'', the choices
are 3am and 3pm, with the former being unlikely. However, if 
someone says ``after 3pm'', there are many choices (e.g., {3:01pm,
3:02pm, ..., 11:59pm}), with those starting at multiples of 30 minutes
being more likely (e.g., {3:30pm, 4pm, ..., 7pm}). Thus, it would be
safer to leave the corresponding entry null. This is how example 
({\exrefnext}) is handled.

\begin{example}
        ({\exref}) & & s1: & ``After the twenty fifth'' \\
        & & ({\exref}, & (null), (august), (after, 25), (null), (null) \\
        & &            & (null), (null), (null), (null), (null)) \\
\end{example}
Although the 31st might be considered a good ending time, there are
other good possibilities (e.g., Thursday 8 September, which is two
weeks afterwards).
Note the special handling of ``after'', in which the day of the week
is not resolved and where the qualifier goes with the day.

\item
{\em Since there is not a separate field for coding duration, omit the
information when it cannot be coded using a pair of starting/ending
fields.}

\begin{example}
        ({\exref}) & & s1: & ``I can meet Thursday for three hours'' \\
        & & ({\exref}, & (thursday), (august), (22), (null), (null), \\
        & &            & (thursday), (august), (22), (null), (null)) \\
 \\
        ({\exref}') & & s1: & ``I can meet Thursday at 1pm for three hours'' \\
        & & ({\exref}', & (thursday), (august), (22), (1), (afternoon), \\
        & &             & (thursday), (august), (22), (4), (afternoon)) \\
\end{example}

\item
{\em Be careful in coding the entries.} Please double-check your
answers. When filling out the form, you might have accidentally copied
over a value from a previous entry.

\end{enumerate}

{
\twocolumn[Appendix: Complete Example of Coding a Scheduling Dialog]
\tiny
\begin{verbatim}
/*
   ;; Dialog Date: 5 March 1993
   ;;
   ;;          Mar 1993
   ;;     S  M Tu  W Th  F  S
   ;;        1  2  3  4  5  6
   ;;     7  8  9 10 11 12 13
   ;;    14 15 16 17 18 19 20
   ;;    21 22 23 24 25 26 27
   ;;    28 29 30 31
   ;;
*/

% [Label, SWeekDay, SMonth, SDay, SHourSpec, STimeOfDay,
%         EWeekDay, EMonth, EDay, EHourEpec, ETimeOfDay]

[

%
%   (1 s1 +beg-sim+ me oye +end-sim+) 
%    (listen to me)
%
[1, [null], [null], [null], [null], [null], 
    [null], [null], [null], [null], [null]],

%   (2 s2 +beg-sim+ hola +end-sim+) 
%    (hello)
%
[2, [null], [null], [null], [null], [null], 
    [null], [null], [null], [null], [null]],

%   (3 s2 hola pepe) 
%    (hello pepe)
%
[3, [null], [null], [null], [null], [null], 
    [null], [null], [null], [null], [null]],

%   (4 s1 hola) 
%    (hello)
%
[4, [null], [null], [null], [null], [null], 
    [null], [null], [null], [null], [null]],

%   (5 s1 que1 tal) 
%    (what+s up)
%
[5, [null], [null], [null], [null], [null], 
    [null], [null], [null], [null], [null]],

%   (6 s2 quieres hacer una reunio1n para la segunda semana de marzo) 
%    (do you want to schedule a meeting for the second week of march)
%
[6, [monday], [march], [8], [null], [null], 
    [friday], [march], [12], [null], [null]],

%   (7 s1 bueno) 
%    (ok *or* well *or* good)
%
[7, [null], [null], [null], [null], [null], 
    [null], [null], [null], [null], [null]],

%   (8 s1 vale) 
%    (ok)
%
[8, [null], [null], [null], [null], [null], 
    [null], [null], [null], [null], [null]],

%   (9 s1 si no hay ma1s remedio) 
%    (if there is no choice)
%
[9, [null], [null], [null], [null], [null], 
    [null], [null], [null], [null], [null]],

%   (10 s1 pues vamos +beg-sim+ a por alla1 ver) 
%    (well we will meet there)
%
[10, [null], [null], [null], [null], [null], 
     [null], [null], [null], [null], [null]],

%   (11 s1 mira) 
%    (look)
%
[11, [null], [null], [null], [null], [null], 
     [null], [null], [null], [null], [null]],

%   (12 s1 a ver) 
%    (let+s see)
%
[12, [null], [null], [null], [null], [null], 
     [null], [null], [null], [null], [null]],

%   (13 s1 a ver) 
%    (let+s see)
%
[13, [null], [null], [null], [null], [null], 
     [null], [null], [null], [null], [null]],

%   (14 s1 a ver) 
%    (let+s see)
%
[14, [null], [null], [null], [null], [null], 
     [null], [null], [null], [null], [null]],

%   (15 s1 hoy lo tengo libre de hecho +end-sim+) 
%    (today i have free then in fact)
%
[15, [friday], [march], [5], [null], [null], 
     [friday], [march], [5], [null], [null]],

%   (16 s2 +beg-sim+ +end-sim+ oye) 
%    (listen)
%
[16, [null], [null], [null], [null], [null], 
     [null], [null], [null], [null], [null]],

%   (17 s2 cinco) 
%    (five)
%
[17, [null], [null], [null], [null], [null], 
     [null], [null], [null], [null], [null]],

%   (18 s2 cinco de marzo no) 
%    (fifth of march right)
%
[18, [friday], [march], [5], [null], [null], 
     [friday], [march], [5], [null], [null]],

%   (19 s1 si1) 
%    (yes)
%
[19, [null], [null], [null], [null], [null], 
     [null], [null], [null], [null], [null]],

%   (20 s2 yo tambie1n) 
%    (me too)
%
[20, [null], [null], [null], [null], [null], 
     [null], [null], [null], [null], [null]],

%   (21 s1 todo el di1a) 
%    (all day)
%
[21, [friday], [march], [5], [null], ['all-day'], 
     [friday], [march], [5], [null], ['all-day']],

%   (22 s2 todo el di1a) 
%    (all day)
%
[22, [friday], [march], [5], [null], ['all-day'], 
     [friday], [march], [5], [null], ['all-day']],

%   (23 s1 pues mira) 
%    (well look)
%
[23, [null], [null], [null], [null], [null], 
     [null], [null], [null], [null], [null]],

%   (24 s1 +beg-sim+ ahora) 
%    (now)
%
[24, [null], [null], [null], [null], [null], 
     [null], [null], [null], [null], [null]],
% treated as disfluency

%   (25 s1 +end-sim+ ahora son) 
%   (now is ???)
%
[25, [null], [null], [null], [null], [null], 
     [null], [null], [null], [null], [null]],
% treated as disfluency

%   (26 s1 ahora son las once y diez) 
%    (now it is eleven ten)
%
[26, [friday], [march], [5], ['11:10'], [null], 
     [friday], [march], [5], ['11:10'], [null]],

%   (27 s1 que1 tal < a > a las doce) 
%    (how about twelve)
%
[27, [friday], [march], [5], ['12:00'], [afternoon], 
     [friday], [march], [5], [null], [null]], 

%   (28 s1 doce a dos) 
%    (twelve to two *or* the twelfth to the seconf *or* the twelfth at two)
%
[28, [friday], [march], [5], ['12:00'], [afternoon], 
     [friday], [march], [5], ['2:00'], [afternoon]], 

%   (29 s2 +beg-sim+ pero) 
%    (but)
%
[29, [null], [null], [null], [null], [null], 
     [null], [null], [null], [null], [null]],

%   (30 s2 +end-sim+ mejor despue1s < de > de almorzar) 
%    (better after eating lunch)
%
[30, [friday], [march], [5], [null], [after, lunch], 
     [friday], [march], [5], [null], [null]], 

%   (31 s1 vale) 
%    (ok)
%
[31, [null], [null], [null], [null], [null], 
     [null], [null], [null], [null], [null]],

%   (32 s1 pues a la una) 
%    (well at one)
%
[32, [friday], [march], [5], ['1:00'], [afternoon], 
     [friday], [march], [5], [null], [null]], 

%   (33 s2 vale) 
%    (ok)
%
[33, [null], [null], [null], [null], [null], 
     [null], [null], [null], [null], [null]],

%   (34 s1 de una a tres) 
%    (from one to three)
%
[34, [friday], [march], [5], ['1:00'], [afternoon], 
     [friday], [march], [5], ['3:00'], [afternoon]], 

%   (35 s2 vale) 
%    (ok)
%
[35, [null], [null], [null], [null], [null], 
     [null], [null], [null], [null], [null]],

%   (36 s1 perfecto) 
%    (perfect)
%
[36, [null], [null], [null], [null], [null], 
     [null], [null], [null], [null], [null]],

%   (37 s1 mira que1 fa1cil) 
%    (look how easy)
%
[37, [null], [null], [null], [null], [null], 
     [null], [null], [null], [null], [null]],

%   (38 s2 nos vemos) 
%    (see you)
%
[38, [null], [null], [null], [null], [null], 
     [null], [null], [null], [null], [null]],

%   (39 s1 hasta luego) 
%    (bye)
%
[39, [null], [null], [null], [null], [null], 
     [null], [null], [null], [null], [null]],

%   (40 s2 hasta luego) 
%    (bye)
%
[40, [null], [null], [null], [null], [null], 
     [null], [null], [null], [null], [null]],

%   (41 s1 adio1s) 
%    (good-bye)
%
[41, [null], [null], [null], [null], [null], 
     [null], [null], [null], [null], [null]]

].
\end{verbatim}
}

\end{document}